\documentclass[letterpaper, 10 pt, conference]{template/ieeeconf}
\IEEEoverridecommandlockouts                              

\usepackage{graphicx}
\usepackage{caption, subcaption}
\usepackage{amsmath,amssymb,enumerate}

\usepackage{amsthm}
\usepackage{mathtools}
\usepackage{breqn}
\usepackage[usenames,dvipsnames,svgnames,table, xcdraw]{xcolor}
\usepackage[colorlinks=true,pdfpagemode=UseNone,citecolor=black,linkcolor=black,urlcolor=BrickRed]{hyperref}
\usepackage{algorithm, algpseudocode}

\usepackage{blindtext}
\usepackage{gensymb}
\usepackage{xparse}
\usepackage{bm}
\usepackage{lipsum}
\usepackage{mathrsfs}
\usepackage[mathscr]{euscript}
\usepackage{times}
\usepackage[noadjust]{cite} 
\usepackage{multicol}
\usepackage{amsfonts}
\usepackage[utf8]{inputenc}
\usepackage[T1]{fontenc}
\usepackage{textcomp}
\usepackage{balance}
\usepackage{soul}
\usepackage{multirow}
\usepackage{booktabs}
\usepackage{sidecap}
\usepackage{makecell}
\usepackage{array,float}

\renewcommand{\thefigure}{\arabic{figure}}

\definecolor{free}{rgb}{0.9608, 0.5882, 0.3922}
\definecolor{building}{rgb}{0.9608, 0.9020, 0.3922}
\definecolor{barrier}{rgb}{0.5882, 0.2353, 0.1176}
\definecolor{other}{rgb}{0.7059, 0.1176, 0.3137}
\definecolor{pedestrian}{rgb}{1, 0.3137, 0.3922}
\definecolor{pole}{rgb}{0.1176, 0.1176, 1}
\definecolor{road}{rgb}{0.7843, 0.1569, 1}
\definecolor{ground}{rgb}{0.3529, 0.1176, 0.5882}
\definecolor{sidewalk}{rgb}{1, 0, 1}
\definecolor{vegetation}{rgb}{1, 0.5882, 1}
\definecolor{vehicles}{rgb}{0.2941, 0, 0.2941}
\definecolor{fence}{rgb}{0.1961, 0.4706, 1}
\definecolor{sign}{rgb}{0, 0, 1}

\definecolor{scarColor}{rgb}{0.9608, 0.5882, 0.3922}
\definecolor{sbicycleColor}{rgb}{0.2608, 0.9020, 0.3922}
\definecolor{smotorcycleColor}{rgb}{0.5882, 0.2353, 0.1176}
\definecolor{struckColor}{rgb}{0.7059, 0.1176, 0.3137}
\definecolor{sothervehicleColor}{rgb}{1, 0.3137, 0.3922}
\definecolor{spersonColor}{rgb}{0.1176, 0.1176, 1}
\definecolor{sbicyclistColor}{rgb}{0.7843, 0.1569, 1}
\definecolor{smotorcyclistColor}{rgb}{0.3529, 0.1176, 0.5882}
\definecolor{sroadColor}{rgb}{0.5, 0, 1}
\definecolor{sparkingColor}{rgb}{1, 0.5882, 1}
\definecolor{ssidewalkColor}{rgb}{0.2941, 0, 0.2941}
\definecolor{sothergroundColor}{rgb}{0.2941, 0, 0.6863}
\definecolor{sbuildingColor}{rgb}{0, 0.7843, 1}
\definecolor{sfenceColor}{rgb}{0.1961, 0.4706, 1}
\definecolor{svegetationColor}{rgb}{0, 0.6863 ,0}
\definecolor{strunkColor}{rgb}{0, 0.2353, 0.5294}
\definecolor{sterrainColor}{rgb}{0.3137, 0.9412, 0.5882}
\definecolor{spoleColor}{rgb}{0.5882, 0.7412, 1}
\definecolor{strafficsignColor}{rgb}{0, 0, 1}

\title{\LARGE \bf Modeling Uncertainty in 3D Gaussian Splatting through \\ Continuous Semantic Splatting}

\author{Joey Wilson, Marcelino Almeida, Min Sun, Sachit Mahajan, Maani Ghaffari, Parker Ewen, \\ Omid Ghasemalizadeh, Cheng-Hao Kuo, Arnie Sen%
\thanks{J. Wilson, M. Ghaffari, and P. Ewen are with the University of Michigan, Ann Arbor, MI 48109, USA. {\tt\small{\{wilsoniv,maanigj,pewen\}@umich.edu}}}
\thanks{M. Almeida, S. Mahajan, M. Sun, O. Ghasemalizadeh, C. Kuo, and A. Sen are with Amazon Lab 126, Sunnyvale, CA, 94089, USA. {\tt\small{\{mmalmeid,msachit,minnsun\}@amazon.com}},
{\tt\small{\{ghasemal,chkuo,senarnie\}@amazon.com}}}
}

\begin{document}

\setlength{\abovedisplayskip}{3pt}
\setlength{\belowdisplayskip}{4pt}

\maketitle
\thispagestyle{empty}
\pagestyle{empty}

\begin{abstract}
In this paper, we present a novel algorithm for probabilistically updating and rasterizing semantic maps within 3D Gaussian Splatting (3D-GS). Although previous methods have introduced algorithms which learn to rasterize features in 3D-GS for enhanced scene understanding, 3D-GS can fail without warning which presents a challenge for safety-critical robotic applications. To address this gap, we propose a method which advances the literature of continuous semantic mapping from voxels to ellipsoids, combining the precise structure of 3D-GS with the ability to quantify uncertainty of probabilistic robotic maps. Given a set of images, our algorithm performs a probabilistic semantic update directly on the 3D ellipsoids to obtain an expectation and variance through the use of conjugate priors. We also propose a probabilistic rasterization which returns per-pixel segmentation predictions with quantifiable uncertainty. We compare our method with similar probabilistic voxel-based methods to verify our extension to 3D ellipsoids, and perform ablation studies on uncertainty quantification and temporal smoothing. 

\end{abstract}



\section{Introduction}
In order to plan, robots require a world model which captures geometric detail and higher levels of information about their environment. Although some papers propose mapless navigation \cite{MaplessMP3, MaplessImitation, MaplessRL}, maps are still widely used due to an interpretable world model which temporally adapts as robots explore their surroundings. Depending on the robot application, maps can store different types of information to increase scene understanding. 

For many robotic applications, uncertainty of the map is necessary to ensure safe planning in safety-critical environments. In these situations, robots must understand not only the type and location of objects, but confidence in the predictions as well. Uncertainty can arise from noisy perception networks, sensor noise, and sparse views which can ultimately result in incomplete maps. 

\begin{figure}[t]
    \centering
    \begin{subfigure}{0.23\textwidth}
        \centering
        \includegraphics[width=\textwidth]{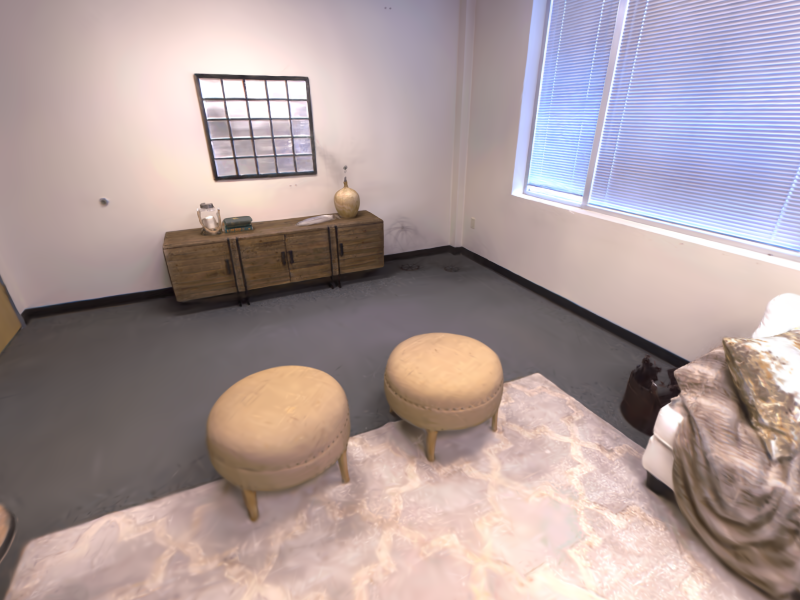}
        \caption{Well fitted 3D-GS render.}
        \label{fig:good_render}
    \end{subfigure}
    \begin{subfigure}{0.23\textwidth}
        \centering
        \includegraphics[width=\textwidth]{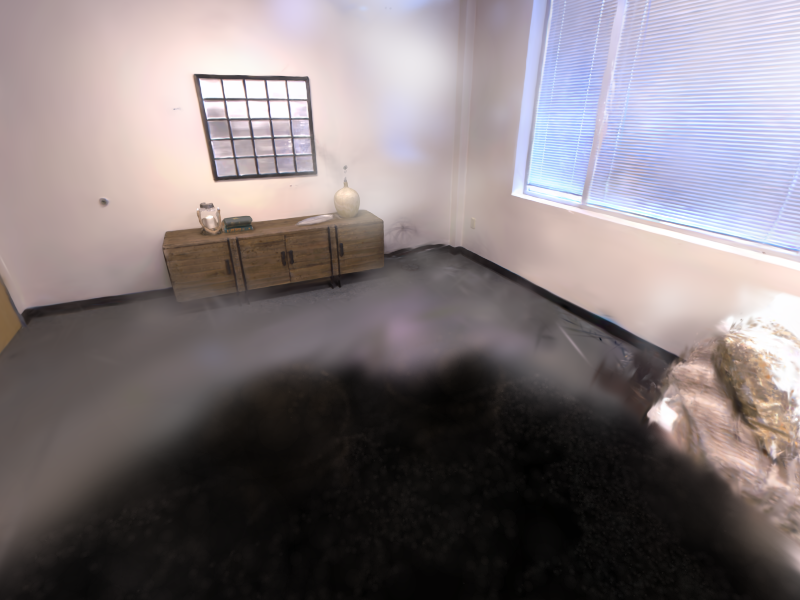}
        \caption{Poorly fitted 3D-GS render.}
        \label{fig:bad_semantic}
    \end{subfigure}
    \begin{subfigure}{0.23\textwidth}
        \centering
        \includegraphics[width=\textwidth]{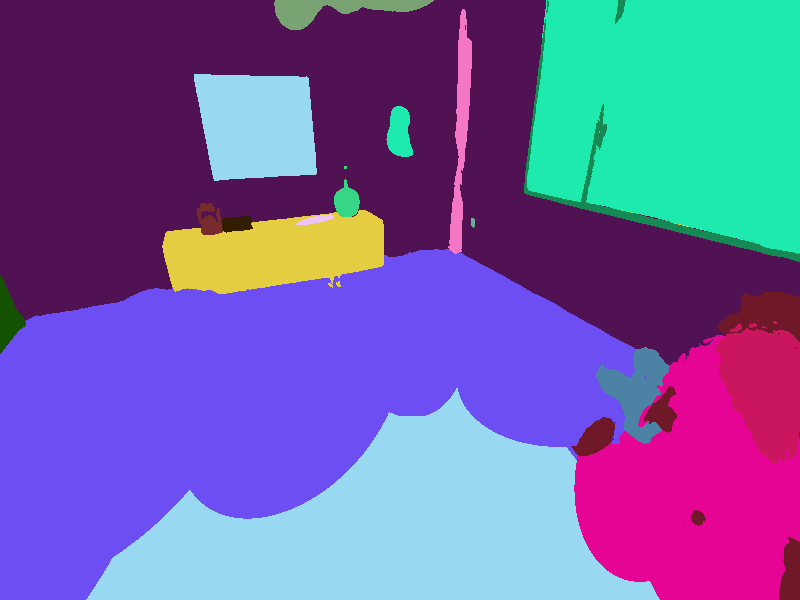}
        \caption{Semantic prediction on poorly fitted render.}
        \label{fig:bad_render}
    \end{subfigure}
    \begin{subfigure}{0.23\textwidth}
        \centering
        \includegraphics[width=\textwidth]{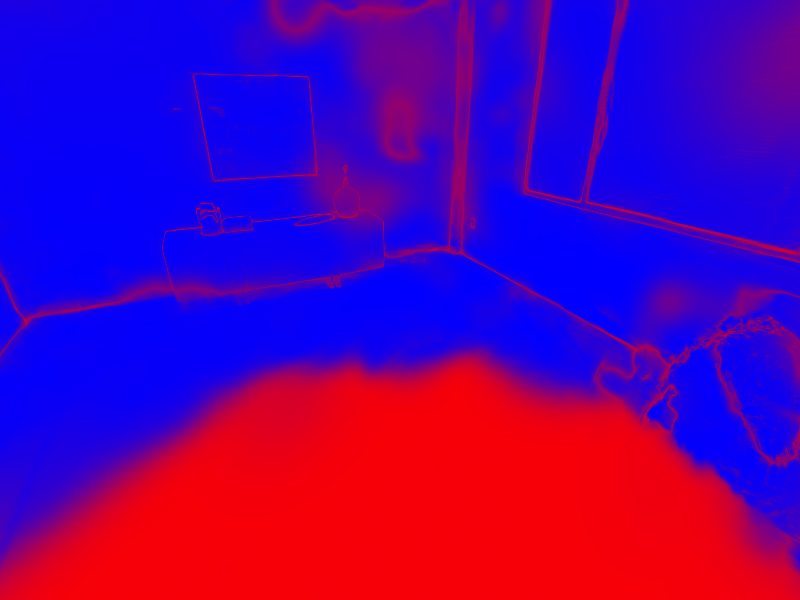}
        \caption{Semantic uncertainty on poorly fitted render.}
        \label{fig:bad_uncertainty}
    \end{subfigure}
    \caption{While 3D-GS may provide high quality renderings of the environment at novel views with sufficient training data, it may fail to render views which are occluded, unseen, or at different angles from the training data. In the above image, CSS produces semantic (c) and RGB (b) predictions at a novel view without sufficient training data, resulting in a blurry render and incorrect segmentation. Through probabilistic inference, CSS identifies blurs and gaps in the render which correlate with reconstruction quality (d). 
    } 
    \label{fig:why_probability}
\vspace{-4mm}
\end{figure}

Continuous mapping combats sparse data by leveraging spatial relations of points to fill in gaps in the map from sparse data probabilistically and with quantifiable uncertainty \cite{BKIProof, BKIOccupancy}. Continuous mapping has been successfully applied to applications such as elevation mapping \cite{ParkerElevation} and semantic mapping, by incorporating measurements into nearby cells in the robotic map through a kernel \cite{MappingSBKI}. The kernel effectively defines the influence of input points over nearby cells probabilistically, leading to a closed form update solution through Bayesian Kernel Inference (BKI). However, one challenge of BKI is defining the kernel function, which is generally hand-crafted and recently was shown to be learnable, resulting in 3D ellipsoid shapes \cite{ConvBKI}. Additionally, BKI has been limited to grid-based solutions which are prone to discretization errors and require accurate depth estimation.

Separately, 3D Gaussian Splatting (3D-GS) proposes a new method for novel view synthesis, which \textit{learns} to model the world explicitly as 3D ellipsoids, with high quality renderings from any angle without the discretization error of grid-based map representations \cite{3D-GS}. 3D-GS has captured the attention of the robotics community, with many methods proposing to add additional features to 3D-GS \cite{LangSplat, Feature3D-GS} and incorporate 3D-GS into simultaneous localization and mapping (SLAM) pipelines \cite{3D-GS_SLAM, 3D-GS_SLAM2}. Some works have recently explored quantifying information gain \cite{FisherRF} or optimal ellipsoid pruning \cite{UncertaintyPruning} in 3D-GS through Fisher Information, however quantifying uncertainty from noisy segmentation networks or novel views remains a challenge. 

In this work, we leverage the insight that 3D-GS learns valid kernels to propose a novel method for uncertainty quantification in 3D-GS. Our method, which we call Continuous Semantic Splatting (CSS), incorporates semantically labelled images in a Bayesian framework to capture the semantic uncertainty of each 3D ellipsoid. Additionally, through a novel rasterization method, we capture the semantic variance from noisy segmentation predictions in pixel space, as well as information on conflicting categories caused by poor renderings at novel views. To summarize, our contributions are:


\begin{enumerate}[i.]
    \item Extend continuous mapping literature from voxel grids to 3D-GS world representation. 
    \item Formulate novel 3D-GS semantic update with quantifiable semantic variance of ellipsoids.
    \item Probabilistic semantic 3D-GS rasterization with quantifiable uncertainty.
\end{enumerate}

\section{Related Work}
In this section we briefly review relevant literature on probabilistic semantic mapping and continuous Bayesian Kernel Inference (BKI), which approaches the ellipsoid world model representation of 3D-GS with quantifiable uncertainty however lacks expressive kernels and remains limited to voxel structures. Next, we present background on the recently developed 3D-GS world model representation, which \textit{learns} an ellipsoid world model representation with rotation yet lacks innate uncertainty quantification. 

\subsection{Probabilistic 3D Semantic Mapping}
In 3D semantic mapping, the goal of the algorithm is to receive sequences of exteroceptive data and update a 3D model of the world with semantic labels and quantifiable uncertainty. A direct approach to solve this problem is to leverage off-the-shelf neural networks to semantically label 3D exteroceptive data, and probabilistically update the map cells occupied by the point through either a voting scheme or Bayesian update \cite{SemanticBayes3D, SemanticFrequent3D, SemanticFusion, SemanticVoting3D, ProbRob}. However, this form of update may lead to sparse maps due to sparse 3D data. To counteract the sparsity of 3D data, Gaussian Processes (GP's) can incorporate input points to form a more complete map \cite{OccupancyGP, GPOM, SemanticGP}. However, GP's have a cubic computational complexity with respect to the number of input points, rendering their use impractical in scenes with high amounts of sensory data. As a result, BKI \cite{BKIProof} is widely used as an approximation of GP's to efficiently model the influence of sparse points on the map through kernel inference \cite{BKIOccupancy}. Semantic BKI \cite{MappingSBKI} extended the BKI framework to semantic labels, and has been applied to applications such as off-road driving \cite{ConvBKIJournal, OffRoadBKI} where uncertainty is critical. 

However, one limitation of Semantic BKI is kernel selection, as kernels are generally hand-crafted. ConvBKI proposes to learn the shape of the kernel through a neural network, resulting in 3D ellipsoid shapes per semantic category \cite{ConvBKI}. While the categorical shapes learned by ConvBKI can enable better informed continuous mapping, the kernels are limited due to the discretization of voxels, the lack of a rotation on the 3D ellipsoid distribution of input points, as well as requiring pre-training on a separate set of semantic segmentation labels. To alleviate these limitations, we propose to extend continuous semantic mapping to 3D-GS, which learns 3D ellipsoid distributions with rotations on more readily available camera data.


\subsection{3D Gaussian Splatting}
3D Gaussian Splatting (3D-GS) \cite{3D-GS} is a new method for novel view synthesis which represents the world explicitly as 3D ellipsoids with color, as opposed to previous methods for novel view synthesis which represented the world implicitly such as Neural Radiance Fields (NeRF's) \cite{NeRF}. Given a set of input images, 3D-GS optimizes the number, location, shape, and color of a set of 3D ellipsoids to best represent the training images. 3D-GS does not require pixel-wise depth predictions of the training images to predict colors, due to a depthwise rasterization process known as alpha compositing. After training, the 3D-GS model can be used to render images at any pose with high quality \cite{3D-GS_Survey}.

Due to the explicit 3D ellipsoid representation and high rasterization quality, 3D-GS has been expanded through works which improve the semantic scene understanding of the model \cite{Feature3D-GS, LangSplat}, and apply the representation to classic robotics problems such as SLAM \cite{3D-GS_SLAM, 3D-GS_SLAM2}. Most works focusing on improving the semantic scene understanding of 3D-GS incorporate features from off-the-shelf segmentation network by learning to render images with features. While this approach has demonstrated success, learning the features does not allow for uncertainty quantification in the model or of the rendered images, and can fail without warning. As previously discussed, uncertainty quantification is an important capacity of robotic maps, which has led several works to examine uncertainty quantification of 3D-GS \cite{UncertaintyPruning, FisherRF}. However, methods for uncertainty quantification of 3D-GS are limited, and extending probabilistic methods from classical robotic mapping to 3D-GS remains an active question. Therefore, we propose to bridge the gap between uncertainty quantification in probabilistic robotic mapping and 3D-GS by leveraging the insight that the 3D ellipsoid shape learned by 3D-GS is a valid kernel which can be incorporated into the BKI framework.

\section{Method}

In this section, we introduce our method Continuous Semantic Splatting (CSS), which probabilistically updates and rasterizes 3D semantic predictions in the 3D-GS representation using continuous BKI. First, we introduce preliminaries on the 3D-GS ellipsoid representation. Next, we introduce preliminaries on BKI and demonstrate how our method extends BKI to 3D-GS. Finally, we present a method for rasterization of the 3D conjugate prior distributions which maintains quantifiable uncertainty in the pixel space. 

\subsection{Preliminaries: Gaussian Splatting} 

3D Gaussian Splatting represents a scene through 3D ellipsoids with location $\mu$, opacity $\alpha$, rotation $R$, scale $S$, and color $c$. Together, the scale and rotation define the shape of the ellipsoid, $\Sigma$, which determine the influence of the ellipsoid over pixels together with the ellipsoid's location and opacity. To render a 2D image, 3D ellipsoids are first converted to a 2D ellipsoid with shape $\Sigma_n'$ and location $\mu_n'$ in a process known as splatting \cite{3D-GS_Survey, 3D-GS}. Given a 2D pixel with location $x'_i$ and a 2D splat $n$, the spatial contribution of the splat is first calculated through a kernel as:

\begin{equation}
    k(x'_i, x'_n) = \text{exp} \left(  -\frac{1}{2}(x'_i - \mu_n')^T {\Sigma'}_n^{-1} (x'_i - \mu_n') \right),
\end{equation}

\noindent where a pixel located at the center of the ellipsoid would result in a contribution of 1. This kernel can be combined with the ellipsoid's opacity to obtain a measure of influence on a passing pixel:

\begin{equation}
    \alpha'_n = \alpha_n \cdot k(x'_i, x'_n).
\end{equation}

\noindent Depths of the 3D ellipsoids are incorporated into the 2D rendering through alpha compositing, which weights the contribution of each ellipsoid to the pixel's color through depthwise sorting as:

\begin{equation}
    \kappa(x'_i, x'_n) = \alpha'_n \prod_{j=1}^{n-1} (1 - \alpha'_j),
\end{equation}

\noindent resulting in a final blended pixel color of:

\begin{equation}
    C_i = \sum_{n=1}^N c_n \kappa(x'_i, x_n').
\end{equation}

Altogether, $\kappa(x'_i, x_n')$ defines the contribution of ellipsoid $n$ to the total color $C$, as a function of the ellipsoid's depth, shape, and opacity. Note that the contribution of the ellipsoid evaluates to a number between 0 and 1 in all cases, with a value of 1 when the pixel ray terminates exactly at the center of the 3D ellipsoid. Based on this insight, we propose that $\kappa(x', \mu_n')$ is a valid kernel that satisfies the constraints of BKI which we present next.

\begin{figure}[t]
    \centering
    \includegraphics[width=0.8\linewidth]{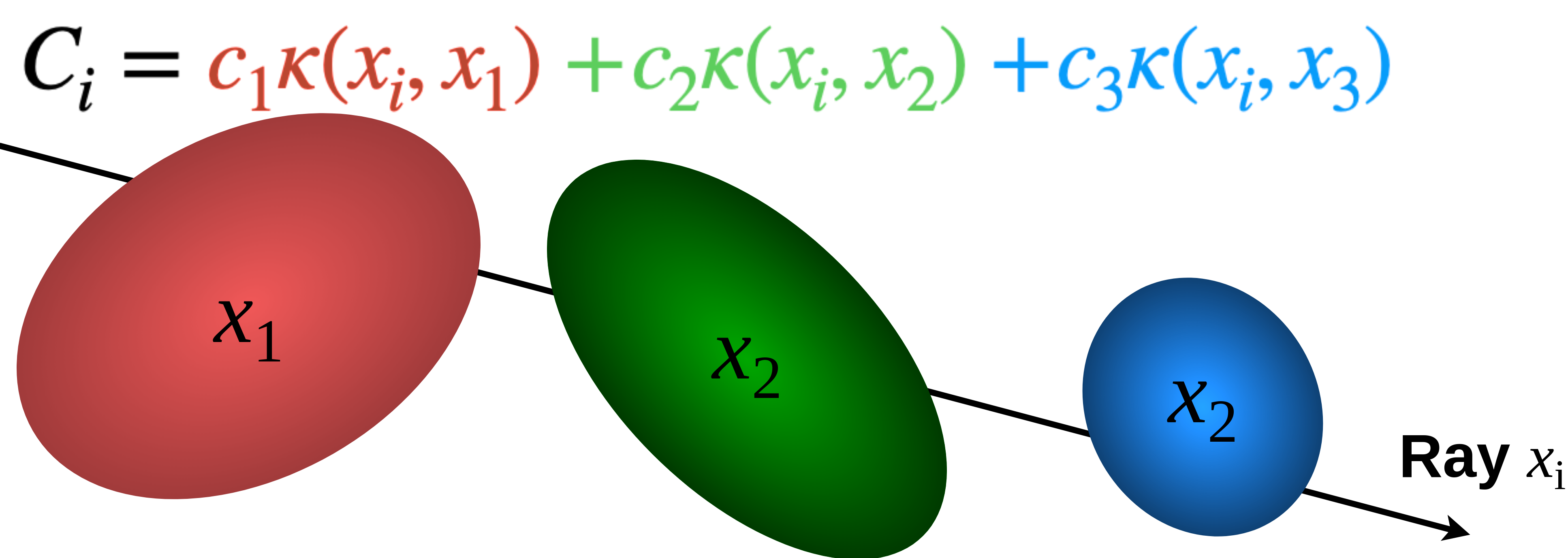}
    \caption{3D-GS renders pixels as a linear combination of 3D ellipsoids, where the influence of each 3D ellipsoid is determined by the spatial position and shape of the ellipsoid $x_n$ relative to pixel $x_i$ as $\kappa(x_i, x_n)$. We propose to leverage the learned expressive kernels of 3D-GS to perform a probabilistic semantic update and rasterization which enables uncertainty quantification. }
    \label{fig:enter-alpha_compositing}
    \vspace{-4mm}
\end{figure}

\subsection{Probabilistic Semantic Update}

Given a fully trained 3D-GS model on a set of images $\mathcal{I}$, we propose to leverage the learned kernels of 3D-GS to perform a probabilistic Bayesian update on the semantic belief of each 3D ellipsoid with BKI. Compared with previous work, our method does not require additional training to learn to render features. Instead, CSS extends classical probabilistic robotic mapping methods to 3D-GS with quantifiable uncertainty. 

For each image in our training set, we first use an off-the-shelf neural network to obtain semantic segmentation predictions $y_i$ for each pixel $x_i$, where $y_i$ is a one-hot encoded vector. Given the training data, our goal is to learn the category of each ellipsoid by defining a categorical likelihood:

\begin{equation}
    p(y_i | \theta_i) = \prod_{c=1}^C (\theta_i^c)^{y_i^c}.
\end{equation}

However, in 3D-GS each pixel is rasterized through partial observations of many ellipsoids. Likewise, the semantic category of each pixel should influence the semantic belief of the same set of ellipsoids. Based on this insight, we propose to perform the Bayesian update with BKI, which relates the likelihood $p(y_i | \theta_i)$ to the extended likelihood $p(y_i | \theta_n, x_i, x_n)$ through a kernel as:

\begin{equation}
    p(y_i | \theta_n, x_i, x_n) \propto p(y_i | \theta_i)^{\kappa(x_i, x_n)}.
\end{equation}

The only requirements on the kernel function are that

\begin{equation}
    0 \leq \kappa(x_i, x_n) \leq 1 \quad \text{and} \quad \kappa(x_i, x_n) = 1 \, \forall \, x_i = x_n,
\end{equation}

\noindent which the 3D-GS kernel satisfies. Based on this observation, BKI provides a method to relate semantic segmentation of pixels to the semantic state of the ellipsoid the pixel passes through. Incorporating the categorical likelihood of a pixel into the extended likelihood formulation yields:

\begin{equation}
    p(y_i | \theta_n, x_i, x_n) \propto \left[ \prod_{c=1}^C (\theta_i^c)^{y_i^c} \right]^{\kappa(x_i, x_n)},
\end{equation}

\noindent which effectively defines the semantic likelihood of a pixel according to the influence of ellipsoid $n$ over the pixel. Next, we define a prior distribution over the semantic state of ellipsoid $n$ using the conjugate prior of the categorical distribution, the Dirichlet distribution. The Dirichlet distribution defines a distribution over a distribution through concentration parameters $\alpha$, as: 

\begin{equation}
    p(\theta_n) \propto \prod_{i=1}^C \theta_{n, c}^{\alpha_n^c - 1}.
\end{equation}

The concentration parameters model the counts of observations of each category, and can be decoded into an expected categorical distribution, and variance. Intuitively, more observations results in lower variance or uncertainty, and the probability of each category can be identified through normalization:

\begin{equation}
    \mathbb{E}[\theta_n^c] = \frac{\alpha_n^c}{\sum_{j=1}^C \alpha_n^j}, \quad \text{Var}[\theta_n^c] = \frac{\mathbb{E}[\theta_n^c] (1 - \mathbb{E}[\theta_n^c])}{1 + \sum_{j=1}^C \alpha_n^j}.
\end{equation}

Combining the conjugate prior distribution and extended likelihood, the Bayesian update over the semantic category of ellipsoid $n$ given all training pixels $\mathcal{D}$ can be written as:

\begin{equation}
    p(\theta_n | x_n, \mathcal{D}) \propto \left[ \prod_{i=1}^N \left[ \prod_{c=1}^C (\theta_n^c)^{y_i^c} \right]^{\kappa(x_n, x_i)} \right]  \prod_{c=1}^C \theta_{n, c}^{\alpha_n^c - 1},
\end{equation}

\noindent which can be simplified to an un-normalized update of the concentrations parameters:

\begin{equation}
    \alpha_n^c \leftarrow \alpha_n^c + \sum_{i=1}^N \kappa(x_i, x_n) y_i^c.
    \label{eq:unnorm_update}
\end{equation}

To summarize, given a set of images $\mathcal{I}$ and a pre-trained 3D-GS model on the set of images, we first label each pixel $x_i$ in the set of training images with an off-the-shelf semantic segmentation network to obtain per-pixel one-hot encoded predictions $y_i$. Next, we adopt an uninformative conjugate prior over the semantic category of all 3D ellipsoids \cite{MappingSBKI}. Last, we update the concentration parameters of each 3D ellipsoid by computing an un-normalized sum over all pixels the ellipsoid influences with Eq. \eqref{eq:unnorm_update}.

\subsection{Probabilistic Semantic Rasterization}

While the BKI update is able to capture variance in the semantic segmentation input network, it does not capture uncertainty in novel views directly. We propose to leverage the rasterization process of 3D-GS to render semantic predictions with uncertainty in the pixel space. Our intuition is that 3D-GS rasterizations can fail in three ways which can be captured through semantic uncertainty. First, semantic rasterization can fail when the input segmentation network is noisy, resulting in high variance predictions. Second, at novel views there may be an absence of ellipsoids, in which case the extended likelihood would have high variance. Finally, novel views may have a blur of objects resulting in a high probability of conflicting categories. Therefore, we propose to model the categorical distribution of the rendered pixel as a linear combination of 3D ellipsoids:

\begin{equation}
    \theta_i =  \sum_{n=1}^N \kappa(x_i, x_n) \theta_n.
\end{equation}

The expectation of the categorical variable can then be rasterized directly through the 3D-GS rasterization process:

\begin{equation}
    \mathbb{E}\left( \theta_i \right) =  \sum_{n=1}^N \kappa(x_i, x_n) \mathbb{E}\left(\theta_n \right),
\end{equation}

\noindent and the variance can be similarly modeled under an assumption of independence between ellipses:
\begin{equation}
    \text{Var}\left( \theta_i \right) =  \sum_{n=1}^N \kappa(x_i, x_n)^2 \text{Var}\left(\theta_n \right).
\end{equation}

When performing alpha compositing to render images, 3D-GS incorporates a background class to fill the absence of ellipsoids. In this case, the background is treated as another ellipsoid, however the weight of the background color is: 

\begin{equation}
    \kappa(x_i, x_{\text{b}}) = 1 - \sum_{n=1}^N \kappa(x_i, x_n),
\end{equation}

\noindent where $x_b$ is the background. Similarly, we propose to incorporate a background distribution, where $\theta_{\text{b}} \thicksim \text{Dir}(\alpha_{\text{b}})$ and $\alpha_{\text{b}}$ is a small positive number uniformly distributed for each category, such that the background has high semantic uncertainty. Due to the formulation of our problem within 3D-GS, variance and expectation both provide important measures of uncertainty with a trade-off in information about conflicting ellipsoid categories (expectation), or few observations (variance).



\subsection{Uncertainty at Image Level}
From the pixels, we may also desire uncertainty at the image level for tasks such as active perception. Inspired by D-Optimality \cite{OptimalExperimentalDesign, OED_Survey}, which defines a functional of the covariance matrix to estimate information, we compute the uncertainty of an image $\mathcal{I}$ from the variance as:

\begin{align}
    U(\mathcal{I_{\text{Var}}}) &= \sqrt[n]{|\bm{\Sigma}_{\mathcal{I}}|} \notag \\
        &= \exp \begin{pmatrix} \frac{1}{n} \sum_{i=1}^n  \log \begin{pmatrix} \text{Var}(\hat{\theta}_i) \end{pmatrix} \end{pmatrix},
\end{align}

\noindent where $n$ is the number of pixels in the image and $\hat{\theta}$ is the categorical variable indexed by the most likely category. We can also obtain a measure of uncertainty from the expectation as:

\begin{equation}
    U(\mathcal{I}_{\mathbb{E}}) = 1 - \frac{\sum_{i=1}^n \mathbb{E}(\hat{\theta}_i)}{n},
\end{equation}

\noindent where the sign is flipped to obtain the uncertainty. This heuristic for uncertainty indicates the pixel-wise average of the probability mass for all non-predicted categories. Intuitively, a low probability mass of non-predicted categories for a pixel indicates low confidence in the predicted categories.
 
\begin{figure*}[t]
    \centering
    \begin{subfigure}{0.32\textwidth}
        \centering
        \includegraphics[width=\textwidth]{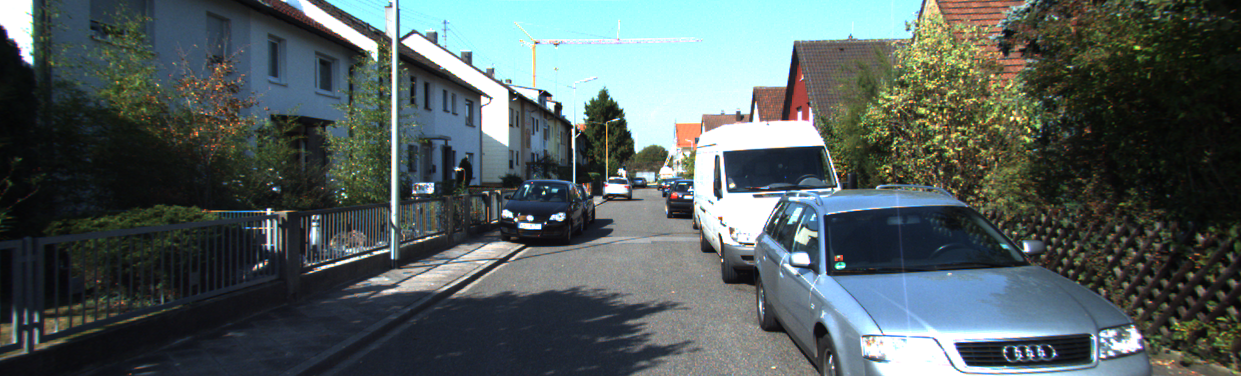}
        \caption{RGB Image of Frame}
        \label{fig:kitti_rgb}
    \end{subfigure}
    \begin{subfigure}{0.32\textwidth}
        \centering
        \includegraphics[width=\textwidth]{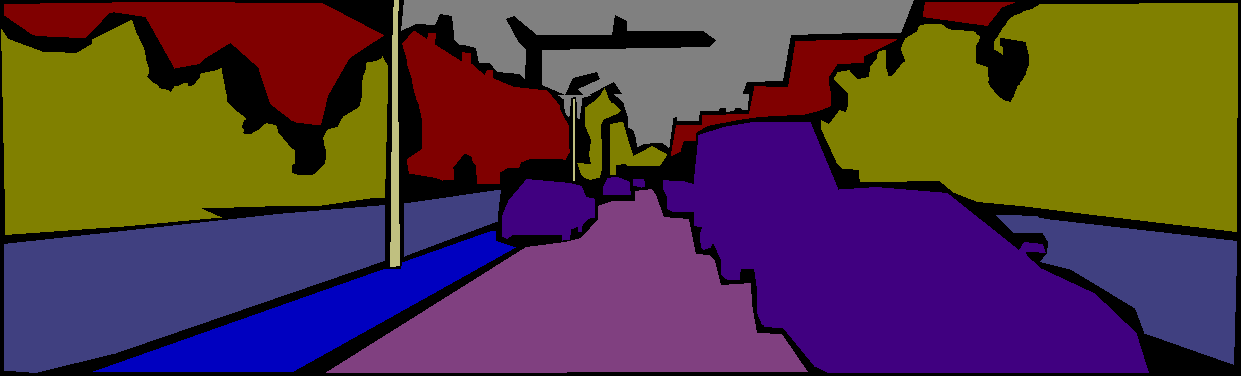}
        \caption{Ground Truth Segmentation.}
        \label{fig:kitti_gt}
    \end{subfigure}
    \begin{subfigure}{0.32\textwidth}
        \centering
        \includegraphics[width=\textwidth]{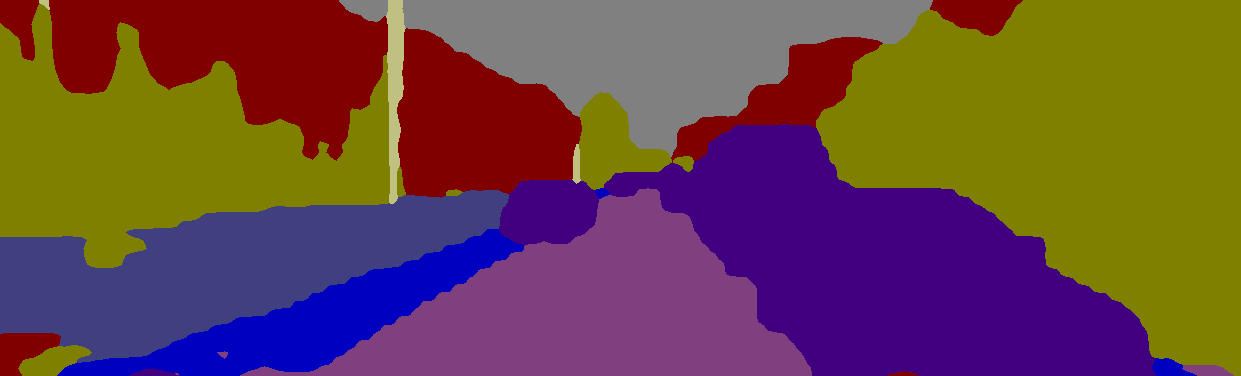}
        \caption{Input Segmentation.}
        \label{fig:kitti_seg}
    \end{subfigure}
    
    \begin{subfigure}{0.32\textwidth}
        \centering
        \includegraphics[width=\textwidth]{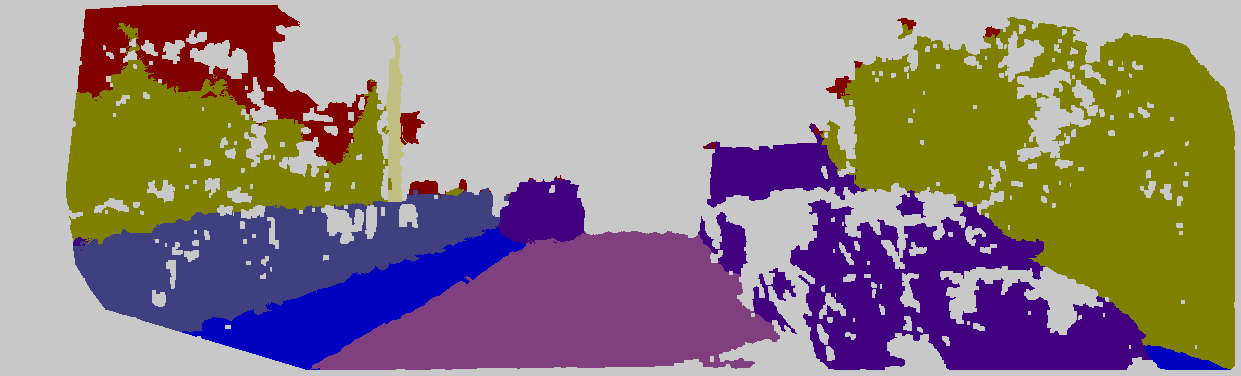}
        \caption{Semantic-BKI Predictions.}
        \label{fig:kitti_voxel}
    \end{subfigure}
    \begin{subfigure}{0.32\textwidth}
        \centering
        \includegraphics[width=\textwidth]{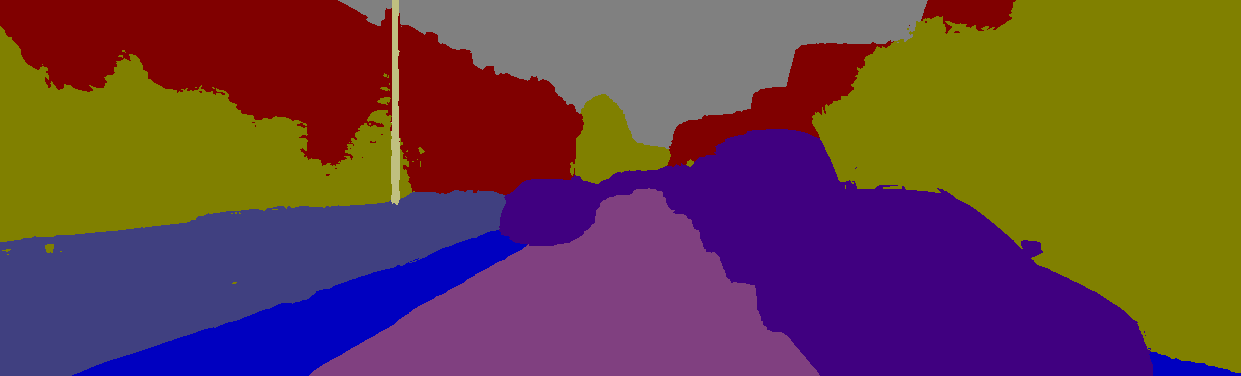}
        \caption{Our Predictions.}
        \label{fig:kitti_3D-GS}
    \end{subfigure}
    \begin{subfigure}{0.32\textwidth}
        \centering
        \includegraphics[width=\textwidth]{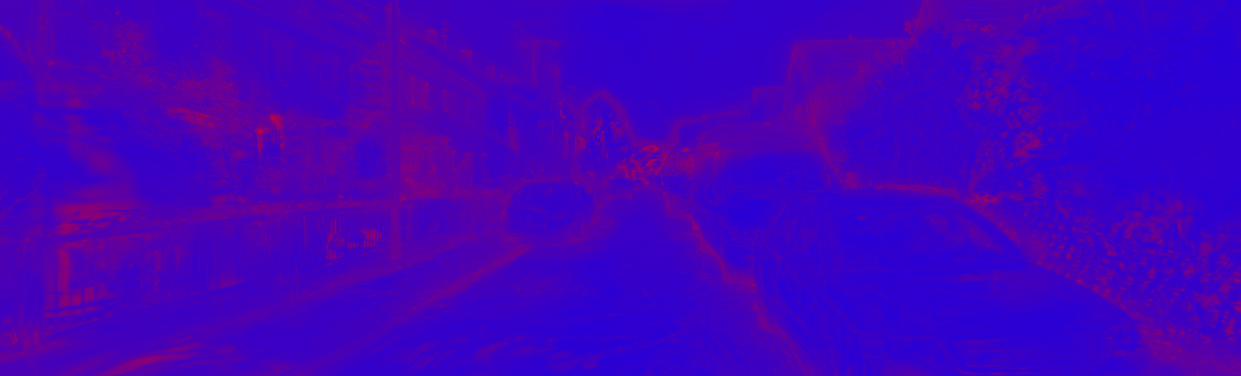}
        \caption{Uncertainty (Expectation) of Our Predictions.}
        \label{fig:kitti_uncertainty}
    \end{subfigure}
    
    \caption{Comparison of our method to a probabilistic voxel baseline on the KITTI driving dataset. Our method achieves similar segmentation results on pixels predicted by the voxel method, and predicts more of the scene due to the lack of a requirement for accurate depth. Additionally, our method does not have discretization, which is beneficial for fine categories such as poles.
    } 
    \label{fig:kitti_comparison}
\vspace{-4mm}
\end{figure*}

\section{Results}
In this section, we verify the probabilistic update and rasterization of our method by comparing to similar voxel-based approaches. We demonstrate that our application of kernel inference to 3D-GS is valid experimentally, as demonstrated by comparable precision, with the benefit of a more complete representation without a requirement of accurate depth. Next, we study the uncertainty quantification capabilities of our method on semantic variance caused by noisy segmentation networks, as well as image rasterization errors caused by insufficient training data. Finally, we study the smoothing effect of our model and the performance gap of our method with perfect segmentation.

\subsection{Comparison to Voxel-Based Methods}
Following the experimental setup of Semantic BKI (S-BKI) \cite{MappingSBKI} which our work is built on, we compare our approach to several probabilistic voxel-based approaches on the KITTI driving dataset \cite{KITTI_Seq_15} to validate our approach against similar mapping algorithms. All methods are provided the same set of images and corresponding semantic segmentation predictions \cite{Deep_Dilated_CNN}. Whereas the voxel-based approaches require pixel-wise depth predictions estimated from ELAS \cite{ELAS_Depth}, our approach requires pre-training to learn the structure of the scene. Models are evaluated on the mean Intersection over Union (mIoU) metric of per-pixel semantic segmentation predictions. The most direct comparisons are Semantic BKI \cite{MappingSBKI}, which applies the BKI operation on voxels with a spherical kernel, and ConvBKI \cite{ConvBKI} which applies the BKI update on voxels with learned per-category kernels.

\begin{table}[b]
    \vspace{-4mm}
    \centering
    \caption{Results on KITTI Odometry sequence 15 \cite{KITTI_Seq_15}. 3D-GS is able to complete more of the scene since it does not rely on accurate depth estimation for training or rasterization.}
    \resizebox{0.48\textwidth}{!}{
    \begin{tabular}{l|cccccccc|c}
        \multicolumn{1}{l}{\bf Method}&
        \cellcolor{building}\rotatebox{90}{\color{white}Building} &
        \cellcolor{road}\rotatebox{90}{\color{white}Road} &
        \cellcolor{vegetation}\rotatebox{90}{\color{white}Vege.} &
        \cellcolor{sidewalk}\rotatebox{90}{\color{white}Sidewalk} &
        \cellcolor{vehicles}\rotatebox{90}{\color{white}Car} &
        \cellcolor{pole}\rotatebox{90}{\color{white}Sign} &
        \cellcolor{barrier}\rotatebox{90}{\color{white}Fence} &
        \cellcolor{pole}\rotatebox{90}{\color{white}Pole} &
        \rotatebox{90}{\bf mIoU (\%)}\\
        \hline
        \vspace{-2mm} \\
        Segmentation \cite{Deep_Dilated_CNN} & 92.1 & 93.9 & \textbf{90.7} & 81.9 & 94.6 & 19.8 & 78.9 & \textbf{49.3} & 75.1\\
        \bottomrule
        \vspace{-2mm} \\
        Yang et al. \cite{YangMethod9} & 32.9 & 85.8 & 59 & 79.3 & 61 & 0.9 & 46.8 & 33.9 & 50\\
        BGKOctoMap-CRF \cite{BKIOccupancy} & 50 & 86.6 & 64.1 & 74.9 & 61 & 0.0 & 47.5 & 36.7 & 52.6\\
        S-CSM \cite{MappingSBKI} & 42.6 & 87.3 & 62.9 & 77.9 & 62.6 & 17.1 & 47.7 & 34.8 & 54.1\\
        S-BKI \cite{MappingSBKI} & 49.3 & 88.8 & 69.1 & 78.2 & 63.6 & 22 & 49.3 & 36.7 & 57.1\\
        \bottomrule
        \vspace{-2mm} \\
        Ours & \textbf{95.5} & \textbf{95.8} & 89.2 & \textbf{84.8} & \textbf{95.5} & \textbf{25.4} & \textbf{80.8} & 45.1 & \textbf{76.5}\\
        
    \end{tabular}
    }
    \label{tab:kitti_iou_all}
\end{table}

First, we compare our approach to voxel-based approaches on all pixels within each image. Qualitative results are shown in Fig. \ref{fig:kitti_comparison}, and quantitative results are shown in Table \ref{tab:kitti_iou_all}. Whereas voxel-based approaches are unable to complete the entire scene due to requiring accurate per-pixel depth, 3D-GS is capable of rendering the entire scene without depth. This difference is visible qualitatively by comparing the gaps in the voxel map generated by Semantic BKI in Fig. \ref{fig:kitti_comparison} to the semantic rendering produced by our method which does not have any gaps. The quantitative evaluation in Table \ref{tab:kitti_iou_all} also supports this claim, as our method achieves a higher mIoU than all probabilistic baselines, as well as improving upon the input segmentation network, highlighting the ability of our method to incorporate sequences of images. While Table \ref{tab:kitti_iou_all} demonstrates improvement in completion, we would like to note that the performance of the voxel-based methods is correlated with the accuracy of depth predictions, and may improve with better depth estimation algorithms. 

\begin{table}[t]
    \centering
    \caption{Results on KITTI Odometry sequence 15 \cite{KITTI_Seq_15} of masked pixels which have predictions from S-BKI.}
    \resizebox{0.48\textwidth}{!}{
    \begin{tabular}{l|cccccccc|c}
        \multicolumn{1}{l}{\bf Method}&
        \cellcolor{building}\rotatebox{90}{\color{white}Building} &
        \cellcolor{road}\rotatebox{90}{\color{white}Road} &
        \cellcolor{vegetation}\rotatebox{90}{\color{white}Vege.} &
        \cellcolor{sidewalk}\rotatebox{90}{\color{white}Sidewalk} &
        \cellcolor{vehicles}\rotatebox{90}{\color{white}Car} &
        \cellcolor{pole}\rotatebox{90}{\color{white}Sign} &
        \cellcolor{barrier}\rotatebox{90}{\color{white}Fence} &
        \cellcolor{pole}\rotatebox{90}{\color{white}Pole} &
        \rotatebox{90}{\bf mIoU (\%)}\\
        \hline
        \vspace{-2mm} \\
        Segmentation \cite{Deep_Dilated_CNN} & 92.1 & 93.9 & 90.7 & 81.9 & 94.6 & 19.8 & 78.9 & 49.3 & 75.1\\
        \bottomrule
        \vspace{-2mm} \\
        Yang et al. \cite{YangMethod9} & 95.6 & 90.4 & \textbf{92.8} & 70.0 & 94.4 & 0.1 & \textbf{84.5} & 49.5 & 72.2\\
        BGKOctoMap-CRF \cite{BKIOccupancy} & 94.7 & 93.8 & 90.2 & 81.1 & 92.9 & 0.0 & 78.0 & 49.7 & 72.5\\
        S-CSM \cite{MappingSBKI} & 94.4 & 95.4 & 90.7 & 84.5 & 95.0 & 22.2 & 79.3 & 51.6 & 76.6\\
        S-BKI \cite{MappingSBKI} & 94.6 & 95.4 & 90.4 & 84.2 & 95.1 & \textbf{27.1} & 79.3 & 51.3 & 77.2\\
        ConvBKI \cite{ConvBKI} & 94.0 & \textbf{95.6} & 91.0 & \textbf{87.2} & 95.1 & 22.8 & 81.9 & 54.3 & \textbf{77.7}\\
        \bottomrule
        \vspace{-2mm} \\
        Ours & \textbf{95.6} & 94.9 & 90.7 & 84.8 & \textbf{95.3} & 8.8 & 79.8 & \textbf{60.8} & 76.3\\
        
    \end{tabular}
    }
    \label{tab:kitti_iou_masked}
    \vspace{-4mm}
\end{table}

Therefore, to understand the ability of our method to incorporate semantic measurements into a 3D model more directly compared to voxel-based methods, we quantitatively compare our model over the same set of masked pixels produced by the probabilistic approaches. Since all algorithms are provided the same set of data, we would expect the results on semantic segmentation produced by the map to be similar. This is visually apparent in Fig. \ref{fig:kitti_comparison}, as well as demonstrated quantitatively in Table \ref{tab:kitti_iou_masked}, where results are mixed between categories. These results confirm our intuition that CSS performs a valid probabilistic update and rasterization. 

\begin{table}[b]
    \vspace{-4mm}
    \centering
    \caption{Results on KITTI Odometry sequence 15 \cite{KITTI_Seq_15} of masked pixels which have predictions from S-BKI. Compared with Table \ref{tab:kitti_iou_masked}, we combine the pole and sign categories into a single pole category.}
    \resizebox{0.48\textwidth}{!}{
    \begin{tabular}{l|ccccccc|c}
        \multicolumn{1}{l}{\bf Method}&
        \cellcolor{building}\rotatebox{90}{\color{white}Building} &
        \cellcolor{road}\rotatebox{90}{\color{white}Road} &
        \cellcolor{vegetation}\rotatebox{90}{\color{white}Vege.} &
        \cellcolor{sidewalk}\rotatebox{90}{\color{white}Sidewalk} &
        \cellcolor{vehicles}\rotatebox{90}{\color{white}Car} &
        \cellcolor{pole}\rotatebox{90}{\color{white}Pole/Sign} &
        \cellcolor{barrier}\rotatebox{90}{\color{white}Fence} &
        \rotatebox{90}{\bf mIoU (\%)}\\
        \hline
        \vspace{-2mm} \\
        S-BKI \cite{MappingSBKI} & 94.5 & \textbf{96.2} & 90.2 & \textbf{87.5} & 95.2 & 50.7 & \textbf{80.1} & 84.9\\
        \bottomrule
        \vspace{-2mm} \\
        Ours & \textbf{95.6} & 94.9 & \textbf{90.7} & 84.8 & \textbf{95.3} & \textbf{61.4} & 79.8 & \textbf{86.1} \\    
    \end{tabular}
    }
    \label{tab:kitti_combined}
\end{table}

One significant difference between our method and voxel-based BKI methods is in the sign and pole categories, where our method achieves a higher mIoU on the pole category but a lower mIoU on the sign category than ConvBKI and Semantic BKI. From examining the confusion matrices of both methods, we find that our method is prone to mislabeling the sign as a pole. This likely occurs because 3D-GS learns to combine visually similar categories into the same structure, whereas the ground truth of the KITTI dataset separates the sign and sign-post as two separate categories. Combining the sign and pole into one pole category, shown in Table \ref{tab:kitti_combined}, we find that our method outperforms Semantic BKI on mIoU, and particularly on the combined pole and sign category. This result demonstrates that 3D-GS is more suitable for complex and detailed environments, where discretization from voxels cannot adequately represent thin objects.

\begin{figure}[t]
    \centering
    \begin{subfigure}{0.23\textwidth}
        \centering
        \includegraphics[width=\textwidth]{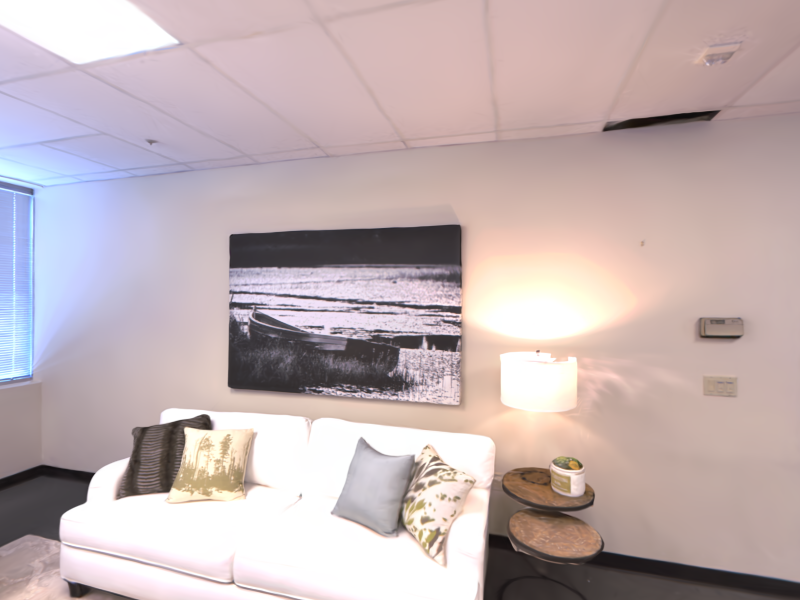}
        \caption{Color rendering. }
        \label{fig:rgb_sem_var}
    \end{subfigure}
    \begin{subfigure}{0.23\textwidth}
        \centering
        \includegraphics[width=\textwidth]{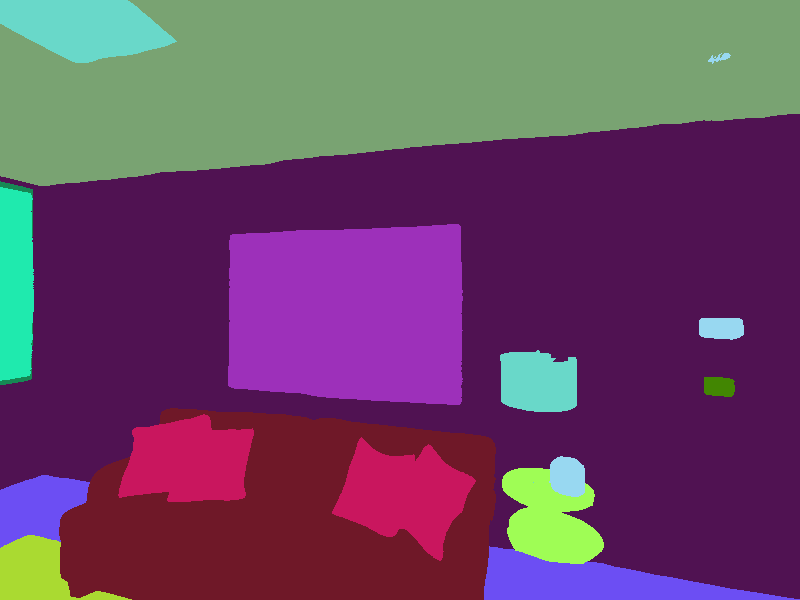}
        \caption{Semantic rendering (GT).}
        \label{fig:rgb_sem_var_gt}
    \end{subfigure}
    \begin{subfigure}{0.23\textwidth}
        \centering
        \includegraphics[width=\textwidth]{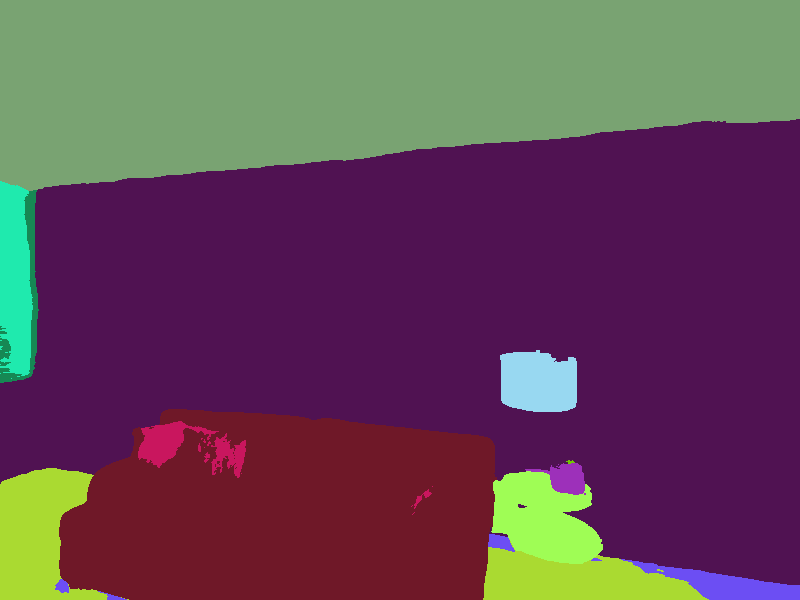}
        \caption{Semantic rendering (LSeg).}
        \label{fig:pred_sem_var}
    \end{subfigure}
    \begin{subfigure}{0.23\textwidth}
        \centering
        \includegraphics[width=\textwidth]{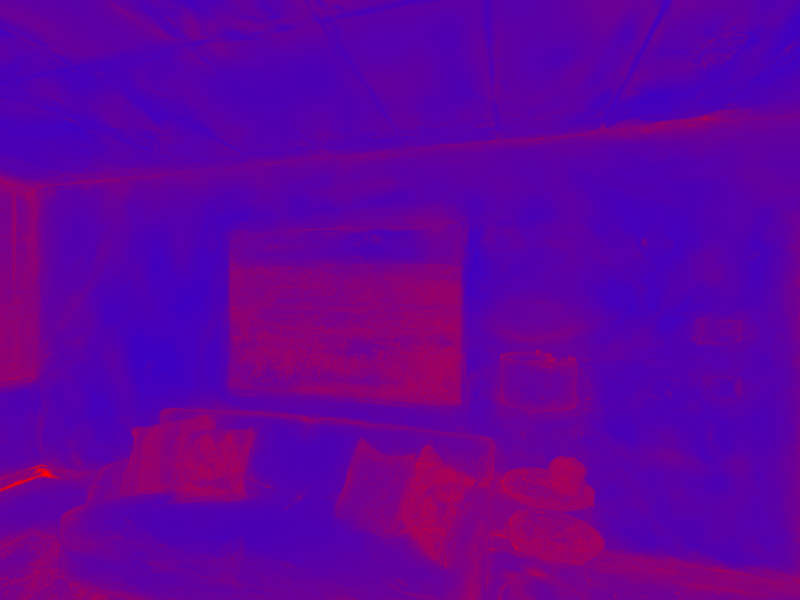}
        \caption{Semantic variance (LSeg).}
        \label{fig:var_sem_var}
    \end{subfigure}
    \caption{Rasterizations from our method on an indoor environment. Our method achieves high quality semantic renderings using ground truth segmentation, shown in (b). Even with noisy segmentation input our method is capable of improving the segmentation with temporal smoothing (c), and can quantify uncertainty (d).
    } 
    \label{fig:sem_var_plot}
\vspace{-4mm}
\end{figure}

\begin{figure}[b]
\vspace{-4mm}
    \centering
    \begin{subfigure}{0.23\textwidth}
        \centering
        \includegraphics[width=\textwidth]{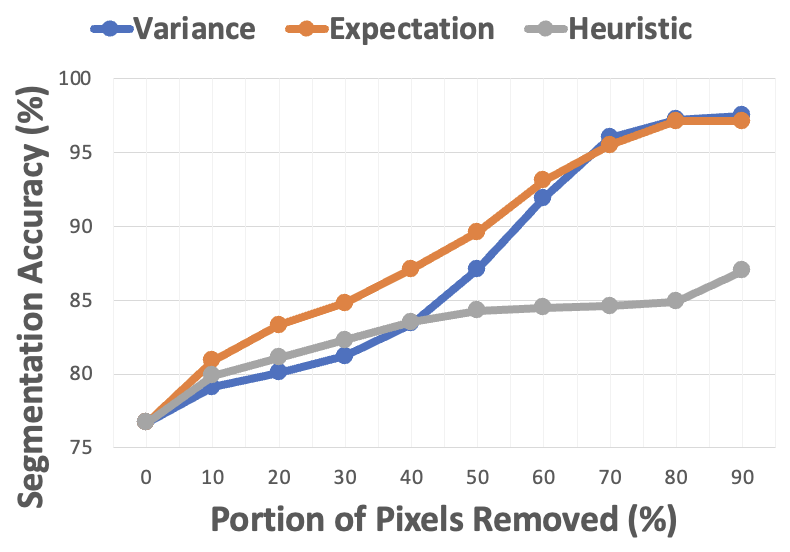}
        \caption{Pixel-level Sparsification.}
        \label{fig:pixel_uncertainty}
    \end{subfigure}
    \begin{subfigure}{0.23\textwidth}
        \centering
        \includegraphics[width=\textwidth]{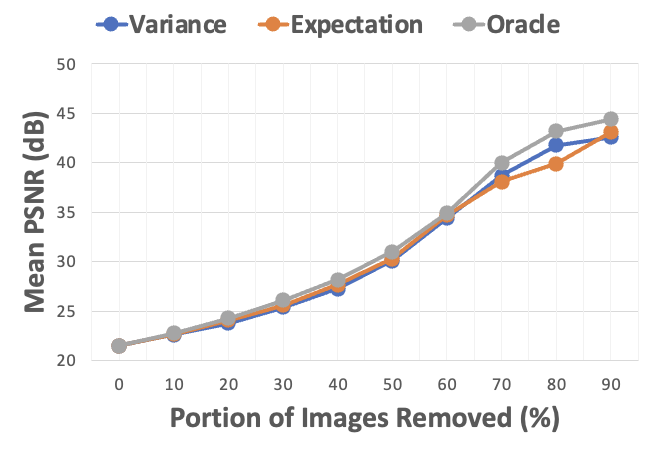}
        \caption{Image-level Sparsification.}
        \label{fig:image_uncertainty}
    \end{subfigure}
    \caption{Sparsification plot of pixel-level and image-level uncertainty. Uncertainty quantification from the expectation and variance are effective at both the image and pixel level.
    } 
    \label{fig:uncertainty_plot}
\end{figure}

\subsection{Uncertainty Quantification}
We perform studies on an indoor environment created with the Replica simulator, which offers high-quality images and ground truth semantic labels \cite{replica, habitat}. In this experiment, 405 images are collected along a manually controlled trajectory in the \textit{room_0} scene in order to simulate a robot path. The data contains 28 unique classes, including basic classes such as wall and floor, as well as more difficult classes such as electrical outlets. An example of the data can be found in Fig. \ref{fig:sem_var_plot}. In order to study uncertainty quantification, we group predictions into bins by confidence, and iteratively remove the least confident sets of predictions, known as sparsification \cite{sparsification}. If uncertainty is properly calibrated, we would expect to see the performance improve as less confident predictions are removed. 

\textbf{Uncertainty of Segmentation} To analyze the ability of our method to quantify uncertainty in the segmentation rasterizations caused by noisy input segmentation networks, we generate segmentation predictions for every image with the open-dictionary segmentation model LSeg \cite{lseg}. Next, we train a 3D-GS model on all images of the dataset and update the concentration parameters of the 3D-GS model with the LSeg  predictions. From the final concentration parameters, we generate pixel-level segmentation predictions for every image, as well as uncertainty calculated by the expectation and variance. Plotting the sparsification in Fig. \ref{fig:uncertainty_plot}, we find that both variance and expectation are highly correlated with segmentation accuracy. Additionally, we compare against a heuristic baseline which computes confidence proportional to the amount of times ellipsoids have been observed. Concretely, pixel-wise confidence is computed as a weighted average of the Dirichlet normalization constants of contributing ellipsoids.

\textbf{Uncertainty of View} Next, we repeat the experiment at the image level, comparing image uncertainty from expectation and variance with the PSNR of the images using the predictions from LSeg. Instead of providing all images from the sequence to train the 3D-GS model and update the concentration parameters, only the first $100$ frames are provided as training data, so that we can examine uncertainty quantification on unseen views. For this experiment, we also compare against an oracle baseline, which quantifies uncertainty according to the actual PSNR of each image. Both methods of quantifying uncertainty are again correlated with image-level PSNR, and with performance close to that of the oracle.

\subsection{Smoothing Effect}

Last, we study the smoothing effect of the continuous mapping operation on the predictions from LSeg in Table \ref{tab:smoothing}. First, we compare the accuracy and mIoU of the predictions from our model with the predictions from LSeg. We find that our method improves over the predictions of LSeg in both metrics, due to temporal incorporation of the segmentation predictions. Next, we repeat the process with ground truth segmentation to examine the loss from using 3D-GS as a map representation. We find that the accuracy is close to $100\%$, however the mIoU suffers in a couple of categories. Similar to before, we find that visually similar classes are blended, such as the electrical outlet being labeled incorrectly as wall. 

\begin{table}[t]
    \centering
    \caption{Segmentation results on Replica dataset.}
    \begin{tabular}{l|c|c}
         Method & mIoU (\%) & Accuracy (\%) \\
         \bottomrule
         LSeg & 25.5 & 70.7 \\
         Ours (LSeg Segmentation) & \textbf{28.4} & \textbf{76.7} \\
         \bottomrule
         Ours (GT Segmentation) & 73.7 & 97.0 \\
    \end{tabular}
    \vspace{-4mm}
    \label{tab:smoothing}
\end{table}
\section{Conclusion}
In this paper, we introduced a novel method for probabilistically incorporating semantic segmentation predictions into a 3D-GS world model. Our method combines the uncertainty quantification abilities of classical robotic mapping methods with the modern ellipsoid representation of 3D-GS by noting that the ellipsoids learned by 3D-GS are a natural extension of continuous mapping. We also proposed novel methods for uncertainty quantification in 3D-GS at the pixel and image level, and found that both the expectation and variance can be useful metrics of uncertainty. For future work, we believe that integrating this method within online 3D-GS frameworks would be a valuable extension for robotic mapping. Additionally, while our proposed method operates on a discrete set of categories, the probabilistic update and rasterization may be extended to open-dictionary continuous splatting through a continuous conjugate prior.

\clearpage
{\small
\balance
\bibliographystyle{IEEEtran}
\bibliography{bib/strings-abrv,bib/ieee-abrv,bib/refs}
}

\end{document}